\pdfoutput=1
\documentclass{article}

\usepackage{longcat_style}
\usepackage[utf8]{inputenc}
\usepackage[T1]{fontenc}
\usepackage{graphicx}
\AtBeginDocument{\setkeys{Gin}{draft=false}}
\usepackage{booktabs}
\usepackage{tabularx}
\usepackage{placeins}
\usepackage{amsmath,amssymb,amsfonts}
\usepackage{microtype}
\usepackage{natbib}
\usepackage{xcolor}
\usepackage[skins,breakable]{tcolorbox}
\usepackage{longcat_cases}
\usepackage{xspace}
\usepackage{url}
\usepackage{hyperref}
\usepackage{fontawesome5}
\usepackage[nameinlink,capitalise]{cleveref}

\definecolor{lcFrontAccent}{HTML}{326A46}

\hypersetup{
  pdftitle={LongCat-DeepResearch Technical Report},
  pdfauthor={Meituan LongCat Team},
  pdfsubject={Technical report on the LongCat-DeepResearch system},
  colorlinks=true,
  linkcolor=blue,
  urlcolor=blue,
  citecolor=blue
}

\newcommand{\longcatdr}{LongCat-DeepResearch\xspace}

\title{LongCat-DeepResearch Technical Report}
\author{Meituan LongCat Team}

\renewcommand{\headeright}{\raisebox{-0.2\height}{%
  \includegraphics[height=1.8em]{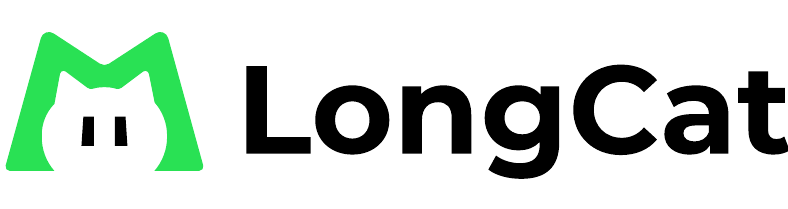}\hspace{0.8em}%
  \includegraphics[height=1.8em]{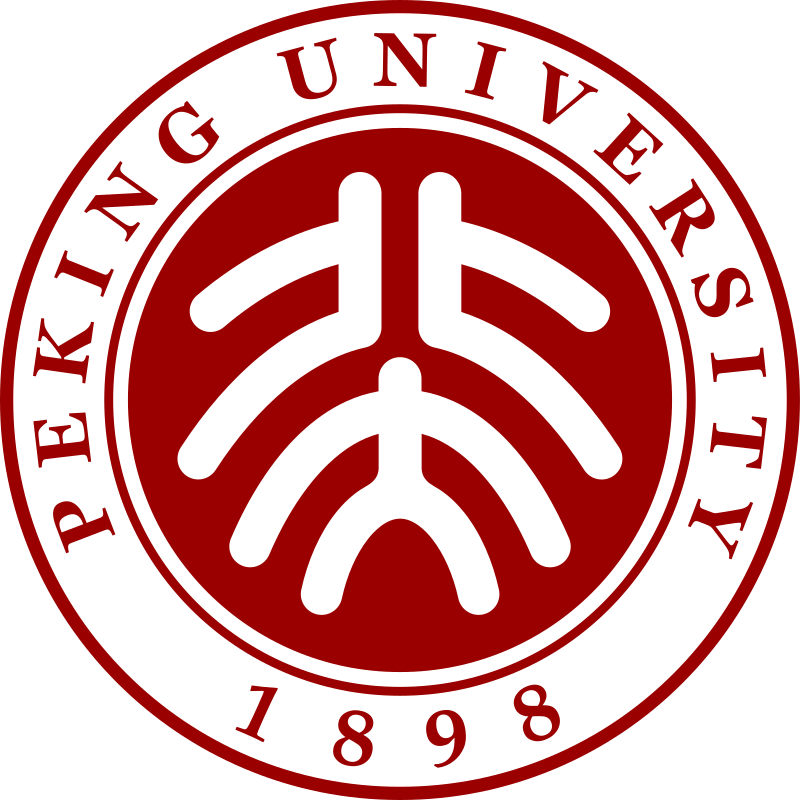}}}
\renewcommand{\shorttitle}{LongCat-DeepResearch}
\AtBeginDocument{\setlength{\headheight}{17pt}}

\begin{document}
\raggedbottom
\maketitle

\begin{abstract}
We present LongCat-DeepResearch, a deep research system that combines an
enhanced LongCat model with a multi-agent workflow for producing comprehensive,
evidence-grounded reports. The workflow separates global planning from detailed
investigation and coordinates revision at the section level. Multiple planning
agents first explore external sources and refine an actionable research plan,
termed ResearchSpec. Research agents then investigate and draft their assigned sections in
parallel, gathering additional evidence in separate contexts as their analyses
develop. Once the sections are assembled, global review guides targeted local
revisions, reducing reliance on repeated full-report rewriting. This workflow
also supports the construction of research tasks and trajectories for the
mid-training and post-training of LongCat's general-purpose models.
LongCat-DeepResearch achieves 55.25 on DeepResearchBench, 51.35 on
DeepResearchBench II, and 79.83 on ResearchRubrics. On an in-house benchmark,
it scores 76.04, ranking second among four compared systems. Development-set
analyses show benefits from combining planning perspectives, while further
planning refinement has mixed effects. Additional editing improves average
automatic readability preference across two benchmarks, with different trends
on each.

\end{abstract}

\begin{center}
\small\color{lcFrontAccent}
\hypersetup{urlcolor=lcFrontAccent}
\href{https://meituan-longcat.github.io/LongCat-DeepResearch/}{\faGlobe\hspace{0.4em}\textbf{Project Page}}
\hspace{1.5em}
\href{https://github.com/meituan-longcat/LongCat-DeepResearch}{\faGithub\hspace{0.4em}\textbf{Code}}
\end{center}

\begin{figure}[!ht]
  \centering
  \includegraphics[width=\linewidth]{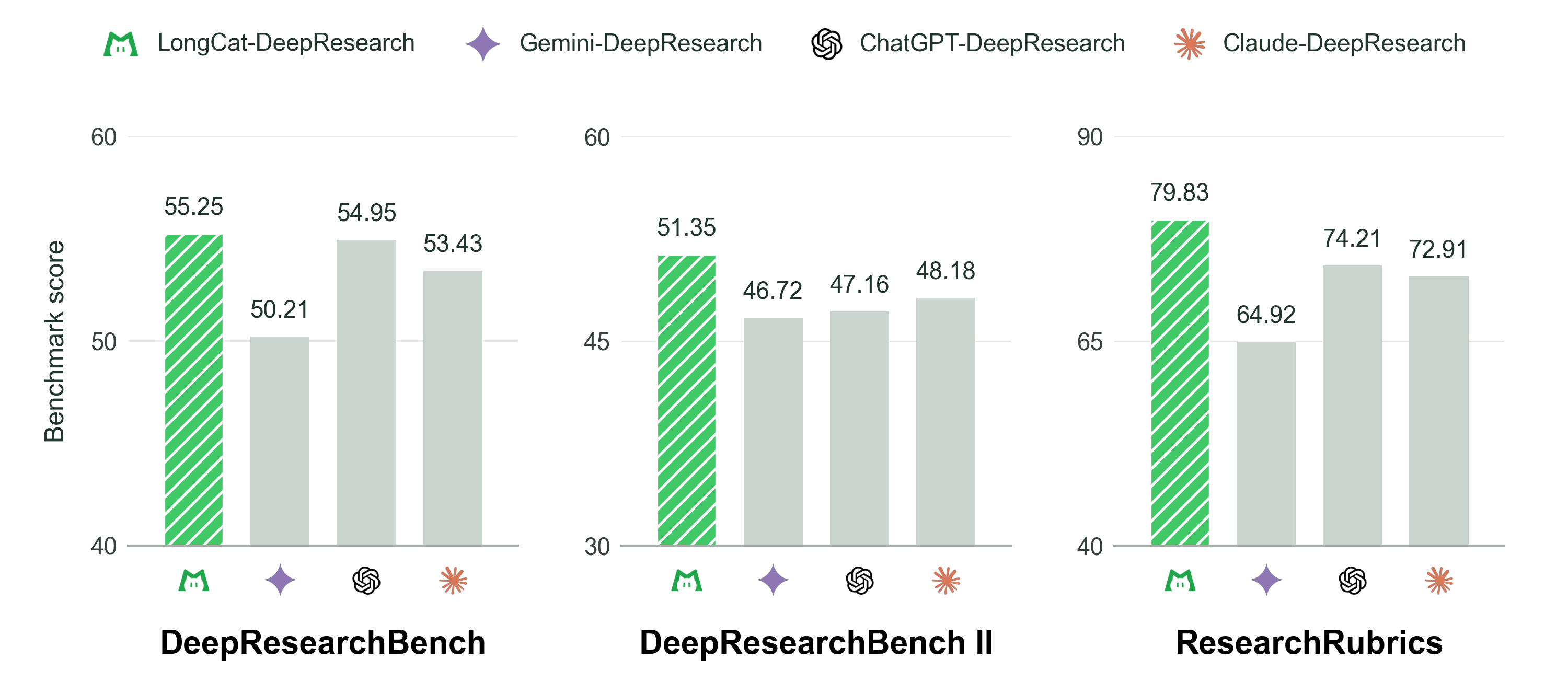}
  \caption{Overall scores on DeepResearchBench, DeepResearchBench II, and
  ResearchRubrics (higher is better). Green bars highlight LongCat-DeepResearch;
  gray bars show the three comparison systems. Evaluation settings and scored
  coverage are detailed in \cref{sec:evaluation} and Appendix~\ref{app:evaluation-details}.}
  \label{fig:benchmark-overview}
\end{figure}

\clearpage
\section{Introduction}
\label{sec:introduction}

Open-ended deep research requires language-model agents to investigate
underspecified questions and gather evidence from diverse external sources
\citep{shao2024storm,chen2024mindsearch,li2025webthinker}. These agents
synthesize their findings into comprehensive long-form reports
\citep{li2025webweaver,zhu2026fsresearcher,li2026agentcpm}.
Unlike conventional information seeking,
the evidence requirements of such reports cannot be fully specified before
the investigation begins. As an analysis develops, emerging explanations
expose missing evidence, unresolved claims require further verification,
and new research questions arise. A research system must therefore establish
a useful initial agenda while allowing further evidence gathering as its
sections develop.
This raises a practical question: \emph{How can deep-research agents
coordinate a shared research agenda while preserving the detailed
investigation needed to write each section?}

Existing systems implement this feedback between research and writing in
different ways. Prior approaches interleave reasoning, retrieval, and drafting
\citep{li2025webthinker}, or use an evolving report to guide subsequent
retrieval and revision \citep{han2025ttddr}. Other approaches organize
investigation through multiple perspectives and evidence-grounded outlines
before detailed composition \citep{shao2024storm,li2025webweaver}.
When this feedback relies on a growing research history or repeated
full-report updates, three difficulties arise. First, evidence, intermediate
reasoning, and draft text compete for finite context; compression may remove
information whose relevance has not yet become apparent. Second, early
findings can anchor the developing narrative and steer subsequent searches,
narrowing the perspectives explored. Third, updating the full report
repeatedly regenerates text beyond the passages affected by new evidence.
The challenge is to retain feedback between research and writing without
concentrating the entire investigation in an expanding draft.

In this work, we introduce LongCat-DeepResearch, a system combining an
improved LongCat model with a research harness centered on an executable
\emph{ResearchSpec}. The key idea is to shift early iteration from the full
report to a compact specification of its research requirements.
Multiple planners independently search and read external sources to develop
candidate specifications. Their proposals are consolidated and refined to
identify missing questions, evidence requirements, and section
responsibilities. This process encodes the emerging global understanding of
the task into ResearchSpec, which serves as a compact proxy for what the
report needs to establish. The downstream synthesis pipeline retains complete section drafts through
assembly and editing. Each researcher receives the complete
ResearchSpec and one assignment, continues gathering evidence in an
independent context, and writes a complete section from its findings. The
resulting section artifacts are assembled into a draft, after which a Global
Editor identifies cross-section issues and Local Editors perform targeted
revisions. This design combines compact global coordination with detailed
local research. ResearchSpec is refined before section dispatch; subsequent
investigation proceeds within the assigned sections, and editing coordinates
their text without reopening the global research agenda.

We evaluate LongCat-DeepResearch on three established benchmarks and an
in-house benchmark. It achieves 55.25 on DeepResearchBench, 51.35 on
DeepResearchBench II, and 79.83 on ResearchRubrics, outperforming the
strongest of the three compared deep-research products by 0.30, 3.17, and
5.62 points, respectively, in the recorded system-level comparison
(\cref{sec:evaluation-protocol}). On our in-house benchmark, LongCat-DeepResearch
ranks second with an overall score of 76.04, exceeding Claude-DeepResearch
at 61.42 and Gemini-DeepResearch at 42.49, and trailing
ChatGPT-DeepResearch's 76.59 by 0.55 points. The same stage interfaces
also support the construction of research questions, task-specific rubrics,
and trajectories used in the mid-training and post-training of LongCat's
general-purpose models.

\section{Related Work}
\label{sec:related-work}

\paragraph{Context management and evidence preservation.}
Long-context access does not ensure reliable use of evidence throughout the
input \citep{liu2023lostmiddle}. MemGPT separates working context from external
storage \citep{packer2023memgpt}, while AgentFold learns fine-grained context
folding and RE-TRAC carries structured summaries across research attempts
\citep{ye2025agentfold,zhu2026retrac}. FoldAct studies the training instability
introduced when generated summaries change future observations
\citep{shao2025foldact}. AdaCoM further shows that the preferred degree of
compression varies with the underlying agent \citep{yi2026adacom}. These works
make compression an explicit, useful operation rather than an incidental
implementation detail. In deep research, SearchSwarm returns compact,
citation-grounded subagent reports to an orchestrator
\citep{lan2026searchswarm}, and Argus synthesizes an answer from a compact
evidence graph \citep{zhang2026argus}. Deep-Reporter maintains a recurrent global
summary and a verbatim local tail during sequential section generation
\citep{ye2026deepreporter}. Our report-synthesis path retains complete section
artifacts through assembly and global-to-local editing, while external
memory and compression address the separate management of research history.

\paragraph{Research planning and structured state.}
WebGPT and ReAct established tool-grounded interaction
\citep{nakano2021webgpt,yao2022react}. STORM develops perspectives before article
writing, and MindSearch decomposes search through an executable graph
\citep{shao2024storm,chen2024mindsearch}. OmniThink uses an information tree
and a conceptual pool to connect knowledge expansion with reflection
\citep{xi2025omnithink}. More recent work makes the research structure
mutable: WebWeaver interleaves evidence acquisition with outline
optimization, AgentCPM-Report alternates drafting and deepening, and
ScaffoldAgent selects outline changes using downstream utility
\citep{li2025webweaver,li2026agentcpm,yang2026scaffoldagent}. Enterprise Deep
Research combines coverage objectives, dependency-guided information sharing,
and evidence-sufficiency conditions \citep{choubey2026dontstop}; SearchOS
externalizes coverage, evidence, pending tasks, and failed searches
\citep{zhang2026searchos}. DecomposeR makes a typed research DAG trainable with
structure-aware rewards \citep{hussain2026decomposer}.

The closest structured-state comparisons also include RhinoInsight, whose
verifiable checklists constrain research actions and whose evidence audit
links sources to drafted content \citep{lei2025rhinoinsight}, and DualGraph,
which maintains separate, co-evolving knowledge and outline graphs
\citep{shi2026dualgraph}. These systems establish structured planning and
evidence binding as existing design choices. In our pipeline, ResearchSpec
defines section-level execution responsibilities that are fixed after planning;
researchers retain their findings in section artifacts passed to editorial
reconciliation. Its role is an execution interface, rather than a persistent
knowledge graph or an input-level specification for personalized query
refinement \citep{yoon2026gsteer}.

\paragraph{Long-form synthesis and revision.}
PaperQA2 develops cited scientific-topic summaries and literature-contradiction
detection \citep{skarlinski2024paperqa2}; OpenScholar combines literature
retrieval, citation-backed generation, and iterative self-feedback
\citep{asai2026openscholar}. These systems establish scientific synthesis as
a research objective beyond isolated fact retrieval. In open-ended report
writing, WebThinker interleaves reasoning, search, and drafting
\citep{li2025webthinker}, while TTD-DR uses an evolving draft to guide retrieval
and iterative report revision \citep{han2025ttddr}. FS-Researcher separates
persistent evidence collection
from multi-session report writing and retains original source files
\citep{zhu2026fsresearcher}. CogGen coordinates a global plan--write--review loop
with section-level work and permits global restructuring
\citep{tian2026coggen}; Ptah maintains inspectable research artifacts and visual
memory for multimodal composition \citep{zhang2026ptah}. Our design assembles the
independently written sections before deciding how to reconcile them: the
Global Editor specifies ownership, while Local Editors revise assigned units.
This scope differs from regenerating the entire document in one invocation,
but does not guarantee that edits retain every useful detail. Mr. Dre
provides direct evidence that report revision can satisfy new feedback while
damaging earlier coverage or citation quality \citep{chen2026mrdre}.
DeepTRACE audits statement-level support and attribution, distinguishing
listed sources from supported claims \citep{venkit2025deeptrace}.
Self-Refine, DuMate, and AREX offer complementary feedback and verification
loops \citep{madaan2023selfrefine,yan2026dumate,lu2026arex}.

\paragraph{Open research systems.}
Open implementations also provide practical precedents for research orchestration.
NVIDIA AI-Q combines structured planning, concurrent researchers, and a
dedicated writer to produce citation-backed reports \citep{nvidia2026aiq}.
LangChain's Open Deep Research supports configurable models and search tools,
with separate stages for compressing findings and writing the final report
\citep{langchain2026opendeepresearch}. GPT Researcher separates planning,
evidence collection, and report synthesis, and supports recursive exploration
\citep{gptresearcher2026system}. Hugging Face's Open Deep Research uses
code-based agents with web-browsing and document-inspection tools
\citep{huggingface2026opendeepresearch}. We build on these established workflow
patterns, focusing on shared planning and the preservation and revision of
complete section drafts.

\paragraph{Synthetic research tasks and trajectories.}
OpenAI's Deep Research System Card describes reinforcement learning on browsing
tasks that include open-ended tasks graded with rubrics
\citep{openai2025drsystemcard}. DR Tulu develops an open long-form research
training approach in which rubrics evolve with the policy and newly acquired
evidence \citep{shao2026drtulu}. These are distinct precedents: rubric-based
supervision predates the evolving-rubric formulation. Tongyi DeepResearch and
Step-DeepResearch describe synthetic tasks and agentic
training across mid-training and post-training
\citep{tongyi2025deepresearch,hu2025stepdeepresearch}. S1-DeepResearch emphasizes
planning, evidence integration, and report generation beyond search-centric QA
\citep{dong2026s1deepresearch}, while Marco DeepResearch emphasizes verification
in task and trajectory construction \citep{zhu2026marco}. For open-ended
reports, requirements must be supported and aligned with the question.
ResearchRubrics contributes expert-written prompts and criteria, including
implicit requirements and negative criteria
\citep{sharma2025researchrubrics}; DeepResearchBench II derives atomic criteria from expert
articles \citep{li2026drb2}. These evaluation resources are distinct from
synthetic training data. DeepRubric constructs evidence trees and jointly
synthesizes queries and rubrics, whereas Quest uses rubric trees to construct
both objective and open-ended research tasks
\citep{zhu2026deeprubric,xie2026quest}. Our data discussion builds on these
principles and describes document provenance, question--rubric alignment, and
trajectory interfaces for research-oriented training data.

\section{Method}
\label{sec:method}

Our method coordinates detailed evidence gathering and report construction
across separate research contexts. \longcatdr assigns section research to
independent contexts and separates global editorial decisions from local
text generation. The resulting intermediate artifacts also provide units
for data construction and stage-wise evaluation.

\subsection{Research Harness}
\label{sec:research-harness}

\begin{figure}[t]
  \centering
  \includegraphics[width=\linewidth]{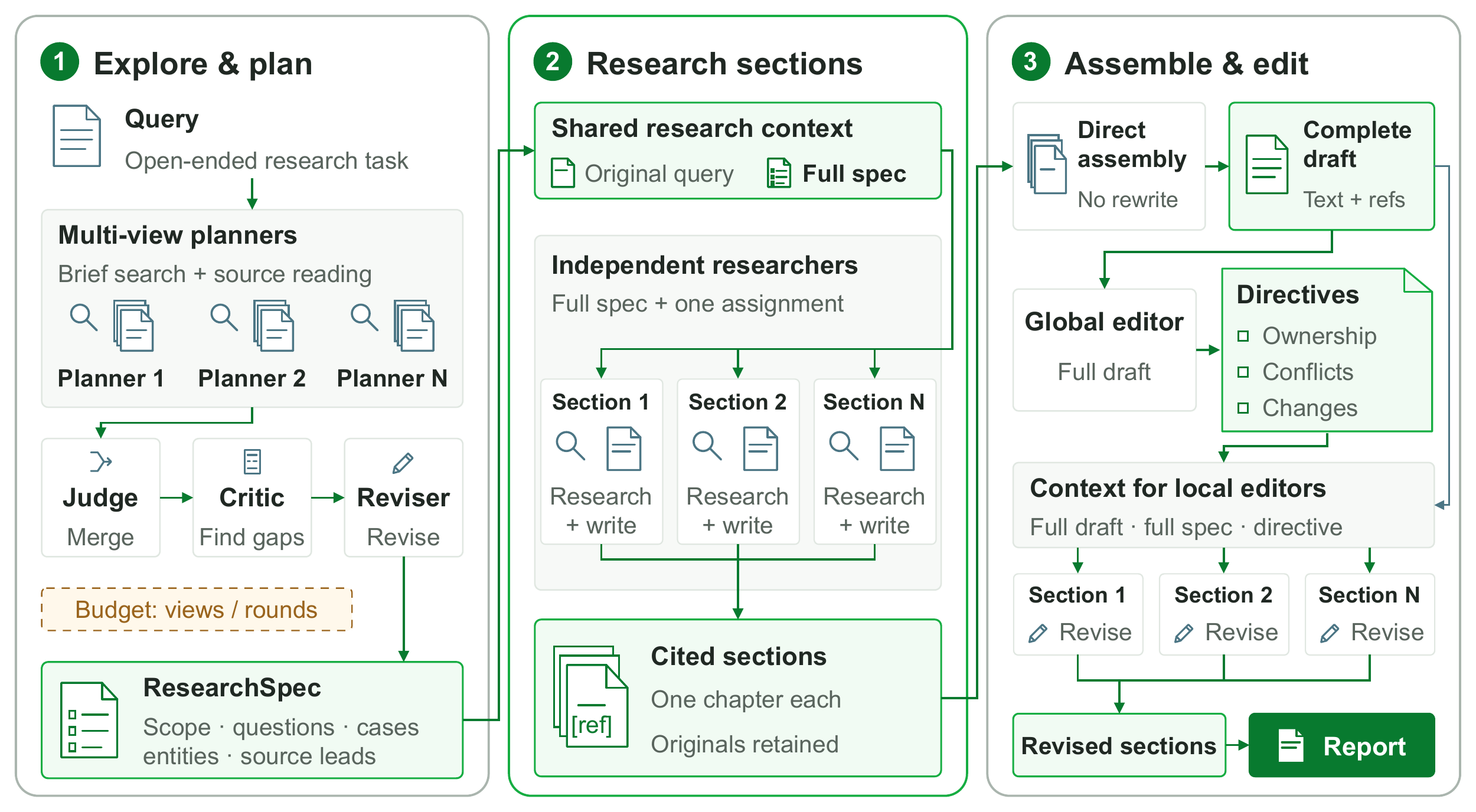}
  \caption{The LongCat-DeepResearch harness. Planning produces a shared
  ResearchSpec; independent Researchers produce complete section drafts;
  global editorial directives guide local revisions after assembly.
  ResearchSpec is fixed during section research, and both editor types
  receive the full assembled draft.}
  \label{fig:research-harness}
\end{figure}

ResearchSpec assigns research responsibilities before detailed evidence
accumulates. Independent Researchers develop their assigned sections, which are then
assembled and edited into a report
(\cref{fig:research-harness}). Retaining these section artifacts does not
imply preserving every retrieved page or every detail of the underlying
interactions.

\paragraph{Explore before committing to a plan.}
\label{sec:planning}
Independent Planning Writers (planners) briefly search and read sources before proposing the
report's structure, grounding the research scope in available evidence beyond
the model's prior knowledge. Their separate explorations can reveal complementary
perspectives, as in perspective-guided article planning
\citep{shao2024storm}. The Planning Judge consolidates the candidate specifications,
the Critic searches for missing questions and cases, and the Reviser incorporates
the feedback. Because section assignments direct subsequent research,
this refinement seeks to identify omissions before a direction is left
uninvestigated. Additional candidates and refinement rounds provide ways to
allocate more computation to planning; their benefit is not guaranteed.

ResearchSpec records each section's scope, research questions, required
entities or cases, and provisional source leads. This compact representation
of the intended report also specifies how research work is divided across
contexts. Its coverage and responsibilities can be inspected and revised
before generating a full draft. Each dispatchable subsection has a unique hierarchical
ID, such as \texttt{S1.2}; we refer to these execution units as sections.
Before dispatch, validators check ID uniqueness, parent relationships,
consecutive ordering, and required fields. Invalid plans undergo repair
or revert to a validated planning candidate; unresolved validation errors
stop dispatch. The findings and source leads in the plan still require
subsequent research and verification. The specification
is revised within the planning stage and then held fixed during section
research; a later research round that reopens the plan is an extension beyond
the current pipeline.

\paragraph{Research and write in independent section contexts.}
\label{sec:synthesis}
Each Researcher receives the original query, the complete ResearchSpec, and
one section assignment. The global specification establishes how its local
work contributes to the report, while separate contexts accommodate the
branches' detailed tool interactions. Researchers receive the shared
specification, without depending on previously generated sections, and can
extend their investigation while addressing the assigned requirements.
They execute concurrently and
write their own citation-bearing sections from the evidence they have acquired.
The same agent therefore carries its local evidence into writing without
first compressing it into a summary for a separate writer. The complete
sections retain the developed arguments and details, reducing the reliance
of subsequent composition on a repeatedly compressed shared history.

\paragraph{Assemble first, then coordinate and edit.}
The harness mechanically assembles sections in ResearchSpec order, retaining
their text and citations even when content overlaps. A Global Editor reads the
complete draft, assigns ownership of repeated material, identifies conflicts,
and specifies changes. Local Editors apply these directives to their assigned
sections using the complete draft and specification as context, drawing on
facts and citations already present in the draft. Their output scope is local,
so one editorial call need not regenerate the entire document. This separates
the global judgment needed for coordination from the generation of revised
text, while keeping the original sections and assembled draft available for
comparison. Both editor types read the full draft and remain subject to input-context
limits. Local output scopes reduce the text regenerated by each call;
repeated full-draft inputs can still incur substantial cost. Research and
editing can omit useful details.

\paragraph{Decoupled interfaces for data construction and evaluation.}
Planning maps the query and explored sources to a ResearchSpec; research maps
section assignments to grounded sections; editing maps a draft and directives
to revised sections. These mappings provide concrete units for targeted data
synthesis and rubric-based evaluation. ResearchSpecs can be checked for coverage before
detailed investigation, sections against their assigned requirements, and
revisions against editorial directives and original text. The decomposition
supports separate improvement and computation allocation at each stage, and provides the foundation for the research-data construction
described next.

\subsection{Research Data Construction}
\label{sec:research-data}
\label{sec:training-longcat}

Data construction follows the harness's stage interfaces: we first build
research tasks with evidence-backed requirements, then collect trajectories
that show how those tasks are planned, investigated, and edited. Independent
source material grounds the questions and rubrics used in this process.

\begin{figure*}[t]
  \centering
  \includegraphics[width=\textwidth]{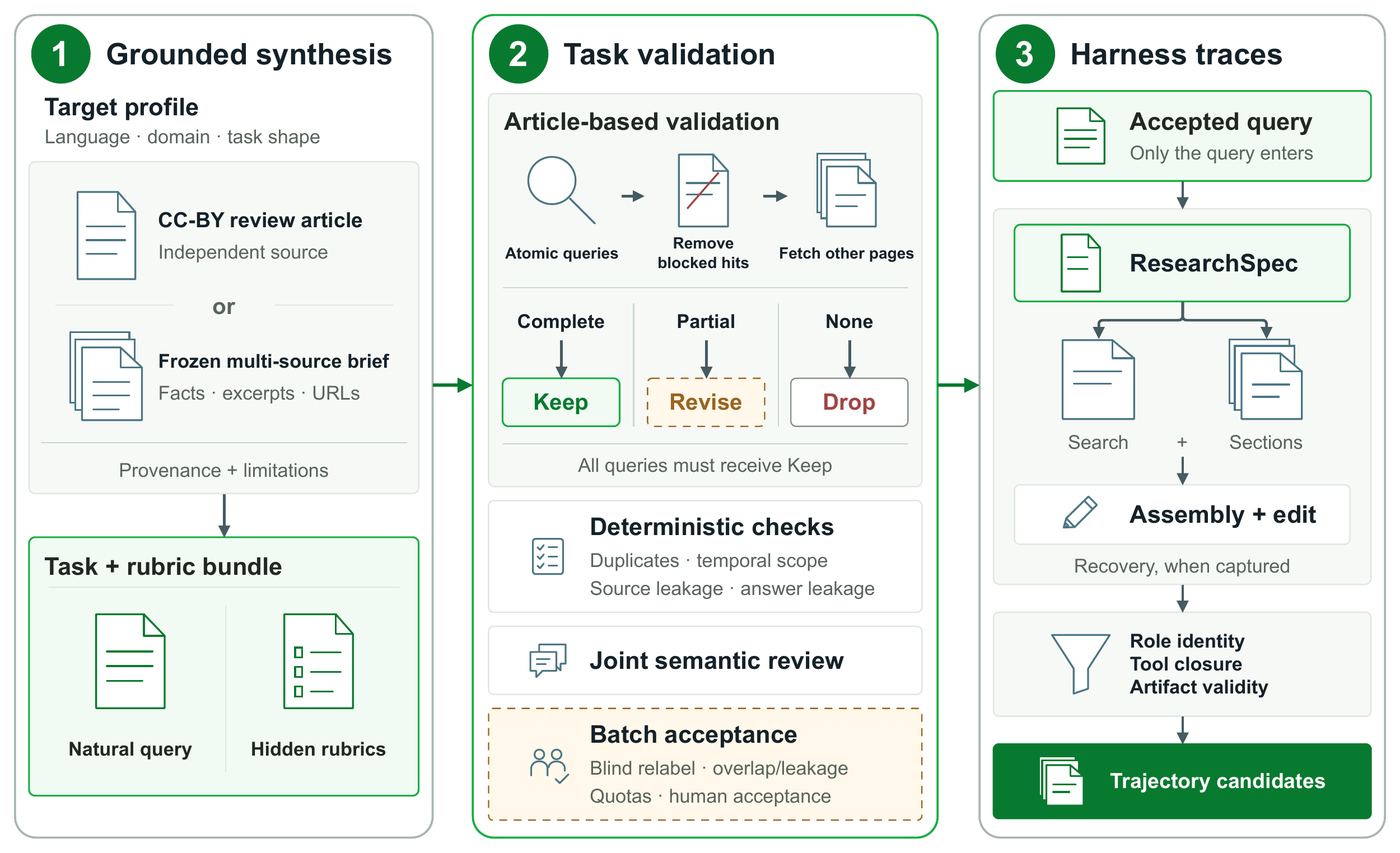}
  \caption{Evidence-grounded construction of research tasks and trajectories. Questions and rubrics are constructed from independent source material; only the query is supplied to the answering agent.}
  \label{fig:data-synthesis}
\end{figure*}

\paragraph{Ground questions and rubrics in source material.}
A target profile specifies language, topic, breadth, and the intended report.
The article-based route starts from an independently licensed review article;
a complementary route uses frozen multi-source briefs containing facts,
excerpts, URLs, and source limitations. Both routes organize evidence for
questions that require explanation or comparison across sources.

Each question is paired with task-specific rubrics. Following evidence-grounded
synthesis principles \citep{zhu2026deeprubric,xie2026quest}, factual criteria
must have supporting evidence, while analytical criteria specify warranted
comparisons or inferences. The rubric can include implicit requirements and
negative conditions, as distinguished in ResearchRubrics
\citep{sharma2025researchrubrics}. Construction evidence and rubrics remain
separate from the query supplied to the answering agent.

\paragraph{Check evidence, searchability, and task quality.}
The article-based route supports a bounded searchability check for atomic
information-recall criteria. It searches for alternative sources, removes hits
from the excluded construction article, fetches the remaining pages, and
checks support for the complete factual requirement. Complete, partial, or
missing support produces \textsc{Keep}, \textsc{Revise}, or \textsc{Drop}; the
gate passes only when all retained criteria receive \textsc{Keep}.
Deterministic checks address duplicate rubrics, answer leakage, excluded-source
leakage, and temporal scope. Joint semantic review checks whether the question,
evidence, and rubric describe a coherent research task.
The bounded search tests evidence availability within its budget, without
certifying exhaustive Web answerability.

\paragraph{Collect and filter stage-specific trajectories.}
Accepted queries drive teacher executions of the harness, recording candidate
and final ResearchSpecs, tool requests and observations, citation-bearing
sections, assembled drafts, and editorial decisions. Parallel branches retain
their actual inputs and dependencies. Filtering checks role identity,
tool-call/response closure, and intermediate-artifact validity, while
preserving distinctions among complete, partial, and failed attempts.
Each record identifies its teacher and harness. Trajectory validity is one
part of dataset selection: dataset-level overlap checks, distribution-aware
selection, and human review where specified are additional checks before
training-data acceptance.

\paragraph{Model training.}
Research-related data are used alongside other data in the mid-training and
post-training of LongCat's general-purpose models to improve deep-research
capabilities. LongCat-DeepResearch combines this model with the research
harness described in \cref{sec:research-harness}.
The system-level scores do not isolate the contribution of this pipeline,
one data source, or one training stage.

\section{Evaluation}
\label{sec:evaluation}

We organize evaluation around four benchmarks: DeepResearchBench, DeepResearchBench II, ResearchRubrics,
and an in-house benchmark. Following each benchmark's official evaluation
setup, DeepResearchBench RACE and DeepResearchBench II are judged by GPT-5.5 (medium), while
ResearchRubrics is judged by Gemini 2.5 Pro. These models serve as
evaluators of the generated reports. Evaluator-specific reference results are available from the
\href{https://huggingface.co/spaces/muset-ai/DeepResearch-Bench-Leaderboard}{official benchmark leaderboard}.
The official snapshots and the four systems evaluated here are presented
together in \cref{app:official-leaderboards}.
The three public benchmarks provide the recorded comparison
with Gemini-DeepResearch, ChatGPT-DeepResearch, and Claude-DeepResearch. The
in-house benchmark is reported in \cref{sec:inhouse}.
Under the evaluation protocol, the generation input is the original task
prompt; benchmark rubrics and reference reports are reserved for scoring.
Public-Web retrieval remains available during generation.

\subsection{Benchmarks}
\label{sec:evaluation-benchmarks}

\paragraph{DeepResearchBench.}
DeepResearchBench contains 100 PhD-level research tasks spanning 22 domains, balanced
between 50 Chinese and 50 English tasks \citep{du2025drb}. Its RACE protocol
assesses each report along four fixed top-level dimensions:
\emph{Comprehensiveness}, \emph{Insight/Depth},
\emph{Instruction-Following}, and \emph{Readability}. RACE generates
task-specific criteria and weights within these dimensions, scores both a
target and a strong reference, and computes the final relative score as
$100\times S_{\mathrm{tgt}}/(S_{\mathrm{tgt}}+S_{\mathrm{ref}})$ on the percentage scale, where $S_{\mathrm{tgt}}$ and $S_{\mathrm{ref}}$ denote the corresponding target and reference scores. Relative scores are computed within each task before macro-averaging across tasks. We report dimension scores where they are available; the current LongCat-DeepResearch
row includes all 100 tasks from the September 14 evaluation. For this
run, we use the benchmark's GPT-5.5 (medium) evaluator branch with its
official prompt, dynamic criteria, reference reports, weights, and aggregation
code unchanged. The reference reports are the April 2025 snapshot generated by
Gemini-DeepResearch. DeepResearchBench
also provides FACT, which extracts
statement--URL pairs, removes duplicate claims associated with the same URL,
and checks whether the retrieved page supports each statement. The official
current pipeline uses GPT-5.4 Mini for extraction and support judgment and Jina
Reader to obtain page text. These quantities should be accompanied by output length because effective-citation
counts are length-sensitive. We do not report FACT results. FACT measures
statement--page support; it does not by itself establish source authority or
the truth of every claim.

\paragraph{DeepResearchBench II.}
DeepResearchBench II contains 132 tasks across the same 22-domain taxonomy, with 66 tasks in
each language \citep{li2026drb2}. The benchmark is grounded in expert-written
investigative articles. The benchmark paper reports 9,430 atomic rubrics constructed through
automatic extraction, self-evaluation, manual revision, and more than 400
hours of expert review. Its rubrics cover \emph{Information Recall},
\emph{Analysis}, and \emph{Presentation}. A rubric passes only when the report
contains the required fact or inference; numerical requirements must be
matched explicitly. We use the official evaluator and report the pass rate for
each dimension together with the official overall score (the mean of
task-level pass fractions across all rubrics). Task rubrics and reference text
are excluded from the actor prompt by this protocol. If a supporting sentence cites a blocked source
article, the corresponding rubric receives the official score of $-1$. The product comparison uses
GPT-5.5 (medium) as the judge.

\paragraph{ResearchRubrics.}
ResearchRubrics pairs 101 realistic prompts with 2,593 expert-written criteria
covering explicit and implicit requirements, synthesis, citation quality,
instruction following, and communication quality
\citep{sharma2025researchrubrics}. We compute a weighted score per task before macro-averaging across tasks.
The LongCat result covers all 101 tasks.
Dimension scores use the sum of score times weight in the numerator and
only positive weights in the denominator; negative-weight criteria contribute
penalties to the numerator. Dimension scores are not averaged to obtain the
overall result. We use Gemini 2.5 Pro as the judge, following the benchmark's
evaluation setup, and retain its original expert-written criteria and weighted
scoring convention. This
benchmark complements DeepResearchBench II by testing adherence to prompt-specific criteria
that include both requested content and requirements inferred from task context.

\paragraph{In-house benchmark.}
The in-house comparison evaluates LongCat-DeepResearch,
Gemini-DeepResearch, ChatGPT-DeepResearch, and Claude-DeepResearch on the
in-house benchmark using an automatic evaluator.
The resulting dimension scores and weighted overall results are reported in
\cref{sec:inhouse}.

\subsection{Comparison Systems}
\label{sec:evaluation-baselines}

The comparison includes LongCat-DeepResearch, Gemini-DeepResearch,
ChatGPT-DeepResearch, and Claude-DeepResearch. The latter three are accessed
through their respective official clients using each provider's own Deep
Research system, including its native tools and budgets. The selected client
models are Gemini 3.7 Flash, GPT-5.6 Sol (xhigh), and Claude Opus 5 (xhigh),
respectively. The product names in the result tables refer to these complete
Deep Research systems with the specified client model selections.
These system-level comparisons do not isolate a model or harness component.
The configuration study in \cref{sec:analysis-generation} compares
the previous LongCat release with ReAct/direct writing and the current harness,
and compares the previous release with the current LongCat model under the
current harness. All three configurations use the full benchmarks.

\subsection{Evaluation Protocol}
\label{sec:evaluation-protocol}

The benchmark task sets contain 100 DeepResearchBench tasks,
132 DeepResearchBench II tasks, and 101 ResearchRubrics tasks.
DeepResearchBench and DeepResearchBench II use GPT-5.5 (medium),
and ResearchRubrics uses Gemini 2.5 Pro. Each benchmark's overall score
is computed per task and then averaged over its scored tasks; ResearchRubrics
retains signed criterion weights and a positive-weight denominator.
Scored coverage varies across systems. We compare the systems as deployed,
including their native research tools and budgets. Detailed coverage and
execution records are provided in \cref{app:evaluation-details}.

\FloatBarrier
\subsection{Main Results}
\label{sec:evaluation-results}

\Cref{tab:current-main-results} reports the current LongCat-DeepResearch comparison.
LongCat-DeepResearch obtains 55.25 on DeepResearchBench, 51.35 on DeepResearchBench II, and 79.83 on
ResearchRubrics. Relative to the strongest of the three listed web products,
these are differences of $+0.30$, $+3.17$, and $+5.62$ points, respectively.
LongCat-DeepResearch has the highest overall scores on these three public
benchmarks among the compared systems. These point estimates use the scored
cohorts listed below and do not establish statistical significance.
The dimension scores show where the observed differences occur and where
weaknesses remain.

\begin{table}[t]
  \centering\small
  \setlength{\tabcolsep}{4pt}
  \renewcommand{\arraystretch}{1.08}
  \begin{tabularx}{\linewidth}{Xrrrr}
    \toprule
    \textbf{Metric} & \shortstack{Gemini-\\DeepResearch} & \shortstack{ChatGPT-\\DeepResearch} & \shortstack{Claude-\\DeepResearch} & \shortstack{LongCat-\\DeepResearch} \\
    \midrule
    \multicolumn{5}{l}{\textbf{DeepResearchBench}} \\
    Comprehensiveness & 49.47 & 55.41 & 53.71 & \textbf{56.21} \\
    Insight / Depth & 50.73 & 55.82 & 53.87 & \textbf{56.48} \\
    Instruction following & 50.88 & 54.99 & 53.59 & \textbf{55.18} \\
    Readability & 49.28 & \textbf{51.51} & 51.35 & 49.96 \\
    \textbf{Overall} & 50.21 & 54.95 & 53.43 & \textbf{55.25} \\
    \midrule
    \multicolumn{5}{l}{\textbf{DeepResearchBench II}} \\
    Information recall & 41.75 & 41.97 & 44.37 & \textbf{46.42} \\
    Analysis & 51.28 & 52.26 & 51.15 & \textbf{60.69} \\
    Presentation & 84.81 & \textbf{86.98} & 81.02 & 83.13 \\
    \textbf{Overall} & 46.72 & 47.16 & 48.18 & \textbf{51.35} \\
    \midrule
    \multicolumn{5}{l}{\textbf{ResearchRubrics}} \\
    Explicit requirements & 76.36 & 83.57 & 79.64 & \textbf{86.97} \\
    Implicit requirements & 60.25 & 70.36 & 70.57 & \textbf{79.52} \\
    Instruction following & 62.39 & \textbf{74.47} & 74.22 & 66.27 \\
    Citation quality & 32.71 & 64.43 & \textbf{67.27} & 65.00 \\
    Synthesis & 60.41 & 69.14 & 70.29 & \textbf{78.92} \\
    Communication & 59.64 & \textbf{64.24} & 53.33 & 60.00 \\
    \textbf{Overall} & 64.92 & 74.21 & 72.91 & \textbf{79.83} \\
    \midrule
    \multicolumn{5}{l}{\textbf{In-house}} \\
    Content & 71.66 & 87.10 & 80.83 & \textbf{87.11} \\
Analysis & 18.61 & 66.58 & 44.07 & \textbf{70.74} \\
Presentation & 21.35 & \textbf{74.64} & 53.20 & 52.99 \\
    \textbf{Overall} & 42.49 & \textbf{76.59} & 61.42 & 76.04 \\
    \midrule
    Descriptive average & 51.09 & 63.23 & 58.99 & \textbf{65.62} \\
    \bottomrule
  \end{tabularx}
  \caption{Public and in-house benchmark results with benchmark-specific dimensions; higher is better. In-house scores use automatic evaluation (\cref{sec:inhouse}); Overall is the weighted total. The descriptive average is the unweighted mean of the four Overall scores,
  which use different scoring scales and protocols. The Gemini, ChatGPT,
  and Claude results are obtained through their official clients using each
  provider's own Deep Research system, with Gemini 3.7 Flash, GPT-5.6 Sol
  (xhigh), and Claude Opus 5 (xhigh) selected, respectively.
  See \cref{app:evaluation-details} for evaluation details.}
  \label{tab:current-main-results}
\end{table}
\paragraph{DeepResearchBench.}
LongCat-DeepResearch leads the overall comparison with 55.25, exceeding
ChatGPT-DeepResearch by 0.30 points. It also has the highest scores for
comprehensiveness (56.21), insight (56.48), and instruction following
(55.18). Readability remains weaker: its 49.96 is below ChatGPT-DeepResearch's
51.51 and Claude-DeepResearch's 51.35.

\paragraph{DeepResearchBench II.}
LongCat-DeepResearch scores 51.35, exceeding Claude-DeepResearch by
3.17 points. It leads in information recall (46.42) and analysis (60.69),
while its presentation score of 83.13 trails ChatGPT-DeepResearch's 86.98.
The strongest dimension-level advantage is in analysis: 60.69 versus
52.26 for the next-highest system.

\paragraph{ResearchRubrics.}
LongCat-DeepResearch scores 79.83, exceeding ChatGPT-DeepResearch by
5.62 points. It leads in explicit requirements (86.97), implicit
requirements (79.52), and synthesis (78.92). Instruction following,
citation quality, and communication each remain below at least one
compared system. These category scores describe complementary aspects
of report quality; their simple average does not reproduce the full-rubric total.

The case study in \cref{sec:main-case-study} illustrates how ResearchSpec,
section research, and editing interact in one recorded report.

\subsection{In-house Benchmark}
\label{sec:inhouse}
We compare Gemini-DeepResearch, ChatGPT-DeepResearch, Claude-DeepResearch,
and LongCat-DeepResearch on the in-house benchmark.
The In-house block of \cref{tab:current-main-results} reports
Content (task coverage and evidential support),
Analysis (reasoning and synthesis), and
Presentation (organization and clarity),
together with the weighted overall score.
An automatic evaluator assesses the reports along these dimensions.
LongCat-DeepResearch scores 76.04 overall,
0.55 points below ChatGPT-DeepResearch's 76.59.

\FloatBarrier

\section{Design Analysis}
\label{sec:analysis}

We study component ablations, ResearchSpec refinement, and Editor scaling
on development subsets, and compare model--harness configurations on the
full benchmarks. Protocols are reported separately for each study.
DRB-I, DRB-II, and RR abbreviate DeepResearchBench, DeepResearchBench II,
and ResearchRubrics, respectively.

\subsection{Component Ablations}
\label{sec:analysis-component-ablation}

\Cref{tab:component-ablation} compares the complete pipeline (Full) with
three component interventions for LongCat. Full
uses three Planning Writers, the Planning Judge/Critic/Reviser, parallel
Researchers, and the Editor. One Writer $\rightarrow$ Reviser runs these
two stages in sequence and removes the Planning Judge and Critic; it
therefore changes the planning procedure as well as Writer count.
One whole-report Researcher receives Full's complete ResearchSpec but
uses a single research history to produce the draft. No Editor uses the
exact draft saved before editing in its originating Full run.

All conditions within each displayed column use identical question IDs.
The LongCat DeepResearchBench II comparison uses the common completed
questions across all four arms. Selection and the stopped case are
documented in \cref{app:analysis-sampling,app:component4-protocol}.

\begin{table}[!htbp]
\centering\small
\setlength{\tabcolsep}{7pt}
\renewcommand{\arraystretch}{1.10}
\begin{tabularx}{\linewidth}{@{}Xrrr@{}}
\toprule
\textbf{Configuration} & \textbf{DRB-II} & \textbf{ResearchRubrics} & \textbf{Avg.} \\
\midrule
Full & \textbf{48.68} & \textbf{79.14} & \textbf{63.91} \\
One Writer $\rightarrow$ Reviser & 44.65 & 74.46 & 59.56 \\
One whole-report Researcher & 45.64 & 78.49 & 62.07 \\
No Editor & 48.60 & 78.00 & 63.30 \\
\bottomrule
\end{tabularx}
\caption{LongCat component scores; higher is better. DRB-II denotes DeepResearchBench II,
and $\rightarrow$ denotes stage order. Avg. is the unweighted mean of the two displayed benchmark scores. Bold marks the highest mean per column.
Detailed configurations are given in \cref{app:component4-protocol}.}
\label{tab:component-ablation}
\end{table}

Full's advantage over simplified planning and a single whole-report
Researcher supports the division between compact global coordination and
detailed local investigation. ResearchSpec can guide the report's shared
agenda while each section develops its evidence in an independent context.

\subsection{ResearchSpec Refinement}
\label{sec:research-spec-analysis}

We examine how planning aggregation and refinement affect final reports on
a fixed ResearchRubrics development subset with LongCat.
W0 uses a fixed single Writer's ResearchSpec; J0 integrates
three Writers through the Judge, before any Critic/Reviser cycle.
R1 and R2 apply one and two successive Critic/Reviser cycles to J0.
Each stage runs the same downstream Researcher and Editor configuration;
planning compaction is disabled.

\Cref{tab:spec-comparisons} compares planned coverage with final-report
quality; the evaluation protocols are given in
\cref{app:researchspec-coverage-evaluation,app:rr-refinement-protocol}.
Plan aggregation improves final-report quality, while refinement further
improves planned coverage. ResearchSpec thus provides a compact proxy for
early report iteration: the system can refine what the report needs to
establish before expanding it into full prose.

\begin{table}[!ht]
\centering\small
\setlength{\tabcolsep}{9pt}
\begin{tabular}{lrrrr}
\toprule
\textbf{Stage} & \textbf{ResearchSpec} & \multicolumn{3}{c}{\textbf{ResearchRubrics}} \\
\cmidrule(lr){2-2}\cmidrule(lr){3-5}
 & Coverage (\%) & Explicit & Implicit & Total \\
\midrule
W0 & 53.76 & 84.34 & 82.21 & 81.67 \\
J0 & 60.47 & 89.09 & 87.11 & \textbf{85.13} \\
R1 & \textbf{63.14} & \textbf{91.59} & 82.76 & 84.34 \\
R2 & 62.48 & 90.91 & \textbf{87.99} & 84.89 \\
\bottomrule
\end{tabular}
\caption{LongCat planning and report scores on a fixed ResearchRubrics development subset.
W0 is the single-Writer plan, J0 the Judge-integrated plan, and R1/R2 one/two Critic/Reviser cycles.
ResearchSpec coverage evaluates the plan before report generation
(\cref{app:researchspec-coverage-evaluation}).
RR Explicit and Implicit score the corresponding report-rubric categories;
Total uses the full rubric. Bold marks the best stage for each metric.
The protocol is given in \cref{app:rr-refinement-protocol}.}
\label{tab:spec-comparisons}
\end{table}

\subsection{Editor Scaling}
\label{sec:editor-refinement}

We study LongCat Editor scaling on fixed development subsets for each
benchmark, with comparisons matched by question. A0 denotes the unedited
draft. A1 applies the original Editor to A0 once. B1 independently applies
an enhanced Editor to the same A0, adding source excerpts and fact-checking
instructions. B2 edits B1, and B3 edits B2, giving two and three enhanced
rounds in total. A1 is a separate control, not the starting point for B1
(\cref{app:editor-refinement-protocol}).
\Cref{tab:editor-refinement-native} reports native scores and average
readability preference for these conditions. The readability evaluation
protocol is detailed in \cref{app:readability-preference-evaluation}.

\begin{table}[!htbp]
\centering\small
\setlength{\tabcolsep}{7pt}
\renewcommand{\arraystretch}{1.10}
\begin{tabularx}{\linewidth}{@{}X*{3}{>{\raggedleft\arraybackslash}p{0.60in}}>{\raggedleft\arraybackslash}p{0.94in}@{}}
\toprule
\textbf{Configuration} & \multicolumn{3}{c}{\textbf{Native report score}} & \multicolumn{1}{c}{\textbf{Readability}} \\
\cmidrule(lr){2-4}\cmidrule(l){5-5}
 & \textbf{DRB-II} & \textbf{RR} & \textbf{Avg.} & \textbf{Avg. preference} \\
\midrule
A0: No Editor & 55.87 & 85.91 & 70.89 & --- \\
A1: Original Editor (1 round) & 56.04 & 85.66 & 70.85 & 50.00 \\
B1: Enhanced (1 round) & 55.15 & 85.18 & 70.17 & 50.42 \\
B2: Enhanced (2 rounds) & 55.68 & \textbf{86.87} & 71.28 & 53.33 \\
B3: Enhanced (3 rounds) & \textbf{58.38} & 86.00 & \textbf{72.19} & \textbf{53.96} \\
\bottomrule
\end{tabularx}
\caption{LongCat Editor scaling. DRB-II (DeepResearchBench II) and RR (ResearchRubrics)
are native report scores on the development subsets; their Avg. is the unweighted mean of the two displayed scores. Readability is
 the mean automatic preference score across the development subsets against the
same A0 draft; it is a transformed preference margin, not a win rate.
50 denotes parity (\cref{app:readability-preference-evaluation}). A dash marks A0 as the ungraded reference.
Bold marks the highest mean per column, without implying statistical significance.}
\label{tab:editor-refinement-native}
\end{table}

Across enhanced Editor rounds, average readability preference improves;
the final round also preserves or improves the native benchmark scores
relative to the unedited draft. This supports coordinated local revision
as a way to develop independently researched sections into a more coherent
report. Benchmark-specific preferences and paired results are given in
\cref{app:editor-refinement-protocol,app:editor-longcat}.

\subsection{Model and Harness Configurations}
\label{sec:analysis-generation}

We compare three configurations on the full benchmark cohorts.
The first uses the previous LongCat release with bounded ReAct research followed
by a direct report. The second pairs the same previous release with the
current harness. The third
pairs the current LongCat model with the current harness. ReAct uses the same
search and page-reading services and a 65,536-token final-output ceiling, with
thinking enabled.

\Cref{tab:generation-comparison} reports native report quality for the three configurations
and their unweighted averages across the three benchmarks.

\begin{table}[!htbp]
\centering\small
\setlength{\tabcolsep}{5pt}
\renewcommand{\arraystretch}{1.10}
\begin{tabular*}{\linewidth}{@{\extracolsep{\fill}}llrrrr@{}}
\toprule
\textbf{Model} & \textbf{Harness} & \textbf{DRB-I} & \textbf{DRB-II} & \textbf{RR} & \textbf{Avg.} \\
\midrule
LongCat (previous release) & ReAct / direct report & 47.65 & 33.13 & 60.33 & 47.04 \\
LongCat (previous release) & Current harness & 53.58 & 48.68 & 71.90 & 58.05 \\
LongCat (current) & Current harness & \textbf{55.25} & \textbf{51.35} & \textbf{79.83} & \textbf{62.14} \\
\bottomrule
\end{tabular*}
\caption{Report quality across model and harness configurations. Previous release refers to LongCat-2.0. All three configurations are evaluated on the full benchmarks. The last row is LongCat-DeepResearch and reproduces the main results from \cref{tab:current-main-results}. Avg. is the unweighted mean of the three displayed benchmark scores. Bold marks the highest score per column.}
\label{tab:generation-comparison}
\end{table}

The gains from changing the harness with the model fixed show the value
of organizing research around explicit requirements and section-level work.
Further gains with the current model support treating model capability and
research orchestration as complementary parts of the system.

\subsection{Case Study: From ResearchSpec to Report}
\label{sec:main-case-study}

We examine a recorded LongCat trajectory for \emph{AI-enhanced portfolio
management} (DeepResearchBench II 012). The user requests a review of
classical theories, AI/ML applications, multi-criteria decision making,
and portfolio optimization and rebalancing, covering methods through
April 2024. The case exposes two concrete operations: revising the research
agenda before section research, and coordinating overlapping evidence after
the section drafts have been assembled. The excerpts below show the plan,
section draft, editorial directive, and final report from this trajectory.

\subsubsection{Turning the Question into Research Assignments}
\label{sec:portfolio-planning-case}

The merged ResearchSpec decomposes the four requested themes into
19 subsections: ten for classical theories, five for AI/ML applications,
two for multi-criteria methods, and two for optimization and rebalancing.
Each subsection contains \emph{What to Cover}, research questions, required
entities or cases, and source leads. These fields separate the topic of a
section from the evidence that its Researcher must seek. The specification
also records shared boundaries, including the publication cutoff and the
priority given to primary, peer-reviewed sources.

\begin{figure}[!ht]
\centering
\includegraphics[width=\linewidth]{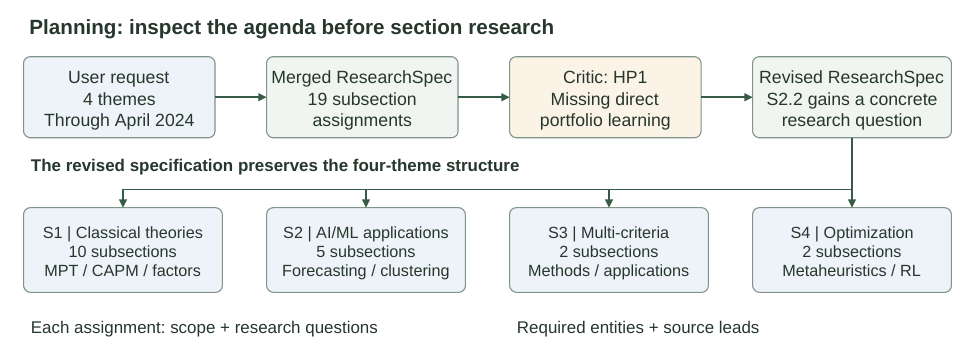}
\caption{Recorded planning artifacts for the portfolio-management case.
The Critic identifies a missing research direction; the Reviser adds it to
S2.2 while retaining the 19-subsection structure. The lower row groups the
resulting assignments by the user's four themes. Selected artifacts are
shown, not every internal model call.}
\label{fig:portfolio-planning-trace}
\end{figure}
\FloatBarrier

\paragraph{A concrete revision before research.}
The merged plan covers forecasting methods but does not name Heaton,
Polson, and Witte or ``Deep Portfolio Theory.'' The Critic flags this
omission as HP1 (the Critic's first listed issue) and directs attention to learning portfolio weights
from data, alongside the existing forecasting agenda. The Reviser adds
the topic to S2.2, including a specific research question and a named entity
for subsequent investigation. The change expands an existing assignment;
it does not create an additional subsection.

\begin{lccase}{\textbf{Planning revision}\hfill\textbf{S2.2 | Time series forecasting}}
\begin{lcband}{CRITIC: IDENTIFY THE MISSING DIRECTION}
\lcquote{HP1: Heaton, Polson \& Witte (2017) --- ``Deep Portfolio Theory''}
\end{lcband}
\begin{lcband}{REVISER: MAKE THE NEW QUESTION EXECUTABLE}
\lcquote{How does Heaton, Polson \& Witte (2017) ``Deep Portfolio Theory'' propose learning portfolio weights directly from data, and how does this differ from the predict-then-optimize approach? \emph{(HP1)}}
\end{lcband}
\end{lccase}

\paragraph{From an assignment to a section artifact.}
The Capital Asset Pricing Model (CAPM) assignment (S1.2) asks which evidence supports or challenges the
model and supplies a survey as a source lead. Its recorded Researcher
output develops that question into a cited section on empirical tests,
Roll's critique, and Fama--French findings. The final report also contains
a dedicated ``Deep Portfolio Theory'' passage under time series forecasting,
preserving the added topic. These are written section artifacts; the next
stage coordinates their overlapping material.

\subsubsection{Coordinating Evidence Across Section Drafts}
\label{sec:portfolio-editing-case}

The Fama--French findings serve two roles in this report. In the CAPM
section they provide an empirical challenge to the model; in the
three-factor-model section (S1.5) they motivate an alternative account of
returns. Independent section writing can therefore repeat a legitimate
piece of evidence. The Global Editor reads the assembled draft and assigns
the detailed treatment to S1.5, while directing S1.2 to retain a brief
mention and a cross-reference. The Local Editor then applies this decision
to the existing CAPM section.

\begin{figure}[!ht]
\centering
\includegraphics[width=\linewidth]{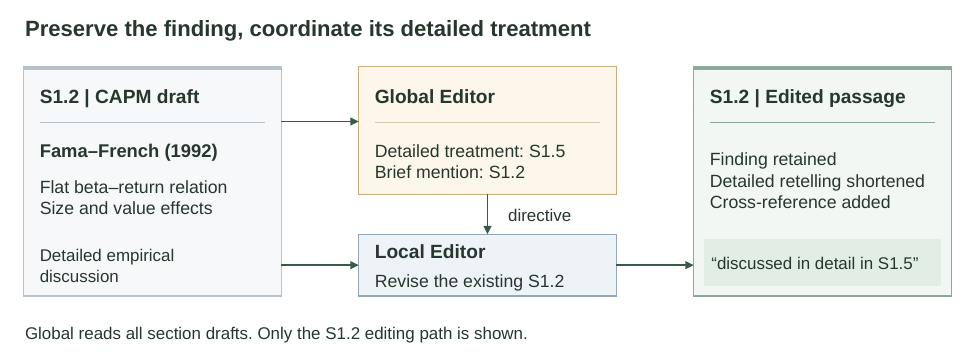}
\caption{The selected CAPM editing path. The Global Editor assigns
cross-section ownership; the Local Editor receives that directive and the
existing section text. The revised passage retains the empirical finding
while referring to S1.5 for its detailed treatment. Other sections in the
Global Editor's assembled context are omitted from this view.}
\label{fig:portfolio-case-trace}
\end{figure}
\FloatBarrier

\begin{lccase}{\textbf{Observed text change}\hfill\textbf{S1.2 | CAPM}}
\begin{lcband}{RESEARCH DRAFT}
\lcquote{Using the Fama-MacBeth methodology on NYSE, AMEX, and NASDAQ stocks (1963--1990), they found that the relation between market beta and average return was \textbf{flat} --- beta had no explanatory power for the cross-section of average returns when used alone.}
\end{lcband}
\begin{lcband}{GLOBAL OWNERSHIP DIRECTIVE}
\lcquote{Fama-French 1992 empirical findings \lcarrow{} S1.5; S1.2 cross-references S1.5}
\end{lcband}
\begin{lcband}{LOCAL REVISION RETAINED IN THE FINAL REPORT}
\lcquote{\emph{Fama-French (1992).} The most damaging empirical challenge to the CAPM came from Fama and French's (1992) finding that market beta had no explanatory power for the cross-section of average returns, while size and book-to-market equity were strongly and robustly related to average returns (discussed in detail in S1.5).}
\end{lcband}
\end{lccase}

\paragraph{Cross-section coordination.}
The plan, draft, directive, and final passage can be linked through the
same section identifier. The revision operates on an existing research
artifact: it preserves the central empirical finding, condenses its local
treatment, and records where the detailed discussion belongs. The Global
plan makes similar ownership decisions for the ``error maximization''
concept and for Non-dominated Sorting Genetic Algorithm II (NSGA-II) details, showing that coordination
concerns the allocation of substantive material across sections, as well
as sentence-level editing. Those directives are recorded decisions; the
CAPM passage above is the specific execution traced here.

\paragraph{Remaining limits.}
The example also exposes a finalization mismatch: the report retains the
internal ``S1.5'' reference while the displayed three-factor-model heading
omits that identifier. The final report still misses some benchmark
requirements, including Treynor and CPPI. Thus, an executable research
agenda and a traceable edit do not guarantee complete coverage or flawless
presentation. Only the final report was scored; the case does not isolate
planning or editing gains. Further excerpts and a contrasting labor-market
case with unresolved coverage gaps appear in \cref{app:case-high,app:case-low}.
\label{sec:portfolio-case-end}

\section{Conclusion}
\label{sec:conclusion}

We presented \longcatdr, a model-and-harness system that shifts early
research iteration from full reports to ResearchSpec. Parallel planning
defines evidence requirements and section responsibilities. Researchers
develop citation-bearing sections independently, and coordinated Editors
revise the assembled draft. These interfaces also support research-task
and trajectory construction.

The system achieves 55.25 on DeepResearchBench, 51.35 on DeepResearchBench II,
and 79.83 on ResearchRubrics. On matched development subsets, the full
LongCat pipeline has the highest component-study means. The full-benchmark
model--harness comparison favors the current system: the current harness
improves scores with the previous release, and the current model further
improves scores under that harness.
Further planning
refinement produces mixed category-level effects, while enhanced Editor
rounds improve average readability preference. These observations are
specific to the evaluated configurations and do not isolate training-stage
contributions.

ResearchSpec coverage supports early inspection of planned requirements;
factual verification and independent report evaluation remain necessary.

\section*{Limitations}
\label{sec:limitations}

\paragraph{Evaluation and automatic assessment.}
Native benchmark scores, planned coverage, and citation support measure
different properties. In the automatic benchmark and component evaluations,
unchanged content can receive different rubric judgments.
Question-bootstrap intervals capture question variation, not uncertainty
from repeated generation or judging. Coverage diagnostics support inspection
of planned requirements; factual verification and independent report evaluation
remain necessary.

\paragraph{Development evidence and reproducibility.}
The component, planning-refinement, and Editor studies use previously
inspected development subsets, some selected using prior model scores. Retrieval windows, realized computation,
and recovery histories vary across studies. Broader task coverage and repeated runs would
help characterize trend stability and resource requirements. Research-task
and trajectory construction belong to a broader model development process.
The model comparisons evaluate the resulting checkpoints; matched training
ablations would help identify contributions from individual data sources and
training stages. Changing web content and hosted research products also limit
exact replay.

\paragraph{Planning and editorial scope.}
ResearchSpec is refined during planning and fixed during section research.
Researchers can investigate further within their assignments, but reopening the
global agenda after drafting remains outside the core pipeline. Complete
section artifacts support assembly and inspection, yet research and editing
can omit relevant details. The framework prioritizes evidence recall and
coverage of research requirements. The resulting reports can be lengthy,
dense, or repetitive, leaving room to improve readability for human readers.
Global and Local Editors read the full draft and
remain subject to input-context limits. Additional refinement has varying
effects across configurations and benchmarks, consuming resources that could
also support evidence gathering. The enhanced Editor jointly changes source
access, instructions, and iteration count, leaving their individual
contributions unresolved.

\section*{Acknowledgments}

We thank the members of the Meituan LongCat Team for their
discussions, feedback, data and evaluation support, and engineering and
infrastructure contributions to this project.

\clearpage
\bibliographystyle{plainnat}
\bibliography{references}

\clearpage
\onecolumn
\appendix
\section{Official Leaderboards and Evaluated Systems}
\label{app:official-leaderboards}

\Cref{tab:official-drb1,tab:official-drb2} place the official GPT-5.5
(medium) leaderboard snapshots accessed on September 14, 2026 alongside
the four systems evaluated in this work. The lower groups reproduce
\cref{tab:current-main-results}; they retain their own system snapshots
and evaluation settings and are not official leaderboard entries.

\begin{table}[!ht]
\centering\small
\setlength{\tabcolsep}{3pt}
\renewcommand{\arraystretch}{1.08}
\begin{tabularx}{\linewidth}{Xrrrrr}
\toprule
\textbf{System} & \textbf{Comp.} & \textbf{Insight} & \textbf{Inst.} & \textbf{Read.} & \textbf{Overall} \\
\midrule
\multicolumn{6}{l}{\textit{Official leaderboard snapshot (September 14, 2026)}} \\
\midrule
ATH-Voicepica-DeepResearch & 56.37 & 56.77 & 55.92 & 53.27 & 55.99 \\
cellcog-max & 56.34 & 57.08 & 55.30 & 51.94 & 55.78 \\
SPIA DeepResearch & 55.26 & 55.99 & 55.12 & 52.70 & 55.12 \\
WhaleCloud-DocChain\_0612 & 55.14 & 55.33 & 54.85 & 52.48 & 54.78 \\
deep-dog-2 & 54.68 & 55.32 & 54.26 & 51.05 & 54.32 \\
bodhi & 54.15 & 54.60 & 54.41 & 51.87 & 54.07 \\
lunon\_full100\_FINAL.submission & 53.42 & 54.83 & 53.41 & 50.48 & 53.51 \\
dalpha-deepresearch & 52.58 & 52.94 & 53.87 & 53.20 & 53.10 \\
sourcery & 50.53 & 52.22 & 51.06 & 49.68 & 51.17 \\
Infosys-MARS-Deepresearch-Agent & 50.96 & 51.79 & 51.77 & 48.62 & 51.12 \\
gemini-2.5-pro-deepresearch & 50.01 & 49.92 & 50.22 & 49.58 & 49.98 \\
openai-deepresearch & 48.05 & 46.69 & 49.29 & 47.62 & 47.84 \\
perplexity-Research & 41.78 & 41.27 & 45.31 & 46.03 & 43.05 \\
grok-deeper-search & 39.65 & 38.12 & 44.62 & 45.72 & 41.22 \\
\midrule
\multicolumn{6}{l}{\textit{Systems evaluated in this work}} \\
\midrule
Gemini-DeepResearch & 49.47 & 50.73 & 50.88 & 49.28 & 50.21 \\
ChatGPT-DeepResearch & 55.41 & 55.82 & 54.99 & 51.51 & 54.95 \\
Claude-DeepResearch & 53.71 & 53.87 & 53.59 & 51.35 & 53.43 \\
\textbf{LongCat-DeepResearch} & 56.21 & 56.48 & 55.18 & 49.96 & \textbf{55.25} \\
\bottomrule
\end{tabularx}
\caption{DeepResearchBench RACE results under GPT-5.5 (medium) evaluation.
Comp., Inst. and Read. denote comprehensiveness, instruction following
and readability. Official rows preserve their published identities;
the lower group uses the system snapshots in \cref{sec:evaluation-baselines}.}
\label{tab:official-drb1}
\end{table}
\FloatBarrier

Scored coverage varies across the systems evaluated in this work
(\cref{app:evaluation-details}). Their web-product snapshots and tool
budgets differ from those of the published entries.

\begin{table}[!ht]
\centering\small
\setlength{\tabcolsep}{3pt}
\renewcommand{\arraystretch}{1.08}
\begin{tabularx}{\linewidth}{Xrrrr}
\toprule
\textbf{System} & \textbf{Info.} & \textbf{Analysis} & \textbf{Pres.} & \textbf{Overall} \\
\midrule
\multicolumn{5}{l}{\textit{Official leaderboard snapshot (September 14, 2026)}} \\
\midrule
GPT-o3 Deep Research & 38.05 & 46.37 & 83.49 & 43.00 \\
Gemini 3 Pro Deep Research & 36.58 & 43.91 & 84.94 & 41.40 \\
Gemini 2.5 Pro Deep Research & 34.28 & 49.11 & 77.98 & 39.98 \\
Doubao Deep Research & 32.58 & 44.12 & 71.53 & 37.36 \\
Perplexity Research & 29.16 & 36.26 & 73.41 & 33.65 \\
\midrule
\multicolumn{5}{l}{\textit{Systems evaluated in this work}} \\
\midrule
Gemini-DeepResearch & 41.75 & 51.28 & 84.81 & 46.72 \\
ChatGPT-DeepResearch & 41.97 & 52.26 & 86.98 & 47.16 \\
Claude-DeepResearch & 44.37 & 51.15 & 81.02 & 48.18 \\
\textbf{LongCat-DeepResearch} & 46.42 & 60.69 & 83.13 & \textbf{51.35} \\
\bottomrule
\end{tabularx}
\caption{DeepResearchBench II results under GPT-5.5 (medium) evaluation.
Info. and Pres. denote information
recall and presentation. The upper group is the
official text-only snapshot; the lower group reproduces our main results.}
\label{tab:official-drb2}
\end{table}
\FloatBarrier

The combined presentation places our system-level comparison alongside
published results without establishing an official leaderboard rank.

\paragraph{Official snapshot provenance.}
The published entries come from
\href{https://huggingface.co/spaces/muset-ai/DeepResearch-Bench-Leaderboard}{the official leaderboard},
Hugging Face Space revision \texttt{ad8fc70b0e6c}, using
\texttt{data\_gpt55/leaderboard.csv} and
\texttt{data\_drb2/leaderboard.csv}. The latter is the GPT-5.5 text-only
reevaluation of public reports, excluding PDF-only submissions and ignoring
embedded DOCX images. The displayed scores retain the dated snapshot values.

\FloatBarrier
\section{Experimental Details and Supplementary Results}
\label{app:evaluation-details}

\subsection{Main Evaluation and Shared Conventions}
The official benchmark task sets are described in
\cref{sec:evaluation-benchmarks}. For ResearchRubrics, LongCat generation uses thinking,
three Planning Writers, twenty maximum tool rounds, and research concurrency
three. Scores are averaged over the reports with completed grades; missing
reports are excluded, and overall and dimension scores use the same evaluated
reports within each system. Scored coverage varies across systems.
ChatGPT-DeepResearch and Gemini-DeepResearch use their common evaluated
ResearchRubrics questions. Claude-DeepResearch's ResearchRubrics overall is
an externally reported aggregate; complete per-task grades are unavailable
for verification in this work.

\paragraph{Web search and page access.}
The harness accesses public web information through internally managed
search and page-reading services. The \texttt{web\_search} interface returns
result titles, URLs, and snippets from the configured search backend;
\texttt{web\_fetch} retrieves and parses selected pages for further reading.
These services provide the retrieval infrastructure used by the research
agents, while the harness controls when to search and which pages to read.

\paragraph{Native scoring and recovery.}
DeepResearchBench uses the benchmark's GPT-5.5 (medium) RACE evaluator.
The development DeepResearchBench II studies use GPT-5.5 (medium),
50-rubric chunks and the complete rubric denominator. ResearchRubrics uses Gemini 2.5 Pro, binary verdicts, signed
criterion weights and the sum of positive weights as denominator.
Recovery retains completed report bytes and grades, reuses exact matching
responses, and charges failed attempts to the original task. Successful
low-score reports are not selectively regenerated. These measurements
include recovery rather than estimating first-attempt reliability.

\paragraph{In-house evaluation.}
\label{app:inhouse-protocol}
The in-house comparison uses an automatic evaluator for report content,
analysis, and presentation. \Cref{sec:inhouse} reports the resulting
evaluation scores.

\subsection{Component Comparison}
\label{app:component4-protocol}
\label{app:analysis-sampling}

The LongCat model is evaluated with thinking enabled.
The comparison uses fixed ResearchRubrics and DeepResearchBench II
development subsets, restricted to questions completed by all four arms.
An incomplete single-Researcher case is excluded from the paired comparison.
The interventions are defined in
\cref{sec:analysis-component-ablation}. The single Researcher's tool
allowance sums the section allowances within the task ceiling, and memory
compression is disabled in the component variants.

The studies have one completed grade per report and do not estimate
repeated-Judge variance.
Official reference-result alignment remains unestablished for these
previously inspected development subsets.

Realized computation and retrieval windows differ across these component
conditions. The reported outcomes are native benchmark scores; dedicated
fact-support and readability evaluations are not complete for these cohorts.

\subsection{Model--Harness Configuration Settings}
All three configurations in \cref{tab:generation-comparison} use the full
benchmarks. ReAct/direct reporting has a
65,536-token final-response ceiling, while the current harness in this
comparison has a 32,000-token per-call ceiling. Cumulative task limits are
600 model calls, 1,048,576 completion tokens, 500 searches and 500 page
reads. These allowances do not imply equal realized compute. Same-model
recovery retains validated plans and sections, original spending and
unchanged scoring rules. Comparisons across retrieval windows and recovered
runs characterize the recorded configurations.

\subsection{ResearchSpec Coverage Evaluation}
\label{app:researchspec-coverage-evaluation}

We evaluate a ResearchSpec before report generation using GPT-5.5 (medium).
The evaluator compares the plan with the same question's rubric and maps
each criterion to supporting plan excerpts and section identifiers;
Researcher and Editor are not executed for this diagnostic. For each
question, coverage is the percentage of positive-weight criteria judged
fully covered, with criteria counted equally. Partial, missing, and
not-plan-evaluable criteria remain in the denominator. We report the mean
of these per-question percentages over the evaluation subset. This score
measures coverage of the full positive-criterion set by the plan, rather
than final-report quality or factual correctness. Including criteria that
cannot be evaluated from a plan limits its interpretation as a measure of
planning completeness; comparisons use the same criterion set. Rubrics and diagnostic judgments are withheld from the
generating agents.

\subsection{Readability Preference Evaluation}
\label{app:readability-preference-evaluation}

GPT-5.5 (medium), with thinking enabled and seed 42, compares each edited
report with its own unedited draft (A0) for organization and expression.
The evaluator reads both full reports under three evaluation guides and
both presentation orders. We orient the margins from both orders toward
the edited report and average them for each question using the evaluator's
guide weighting. The resulting margin $m\in[-2,2]$ is mapped to
$50+25m$: 50 denotes parity, and higher values favor the edited report.
We average the question-level scores within each benchmark, then weight
the two benchmark means equally. Every edited condition is compared
directly with A0; adjacent-round preferences are not accumulated. A0 is
a reference without a self-judgment. These scores express automatic
relative preferences, rather than absolute readability or human ratings.

\subsection{ResearchRubrics Refinement Protocol}
\label{app:rr-refinement-protocol}

\Cref{tab:spec-comparisons} uses the same LongCat development subset
across all four stages. For each question, W0
reuses the first Writer from a single three-Writer planning execution;
it is not selected by report score. J0 is the Judge's integrated plan
before critique or revision. R1 revises J0 once, and R2 revises R1 once
more. The experiment disables planning compaction and explicitly executes
both Critic/Reviser cycles. Each of the four immutable planning snapshots
is passed to the original parallel Researchers and Editor. Generation
uses thinking and seed 42 with the original cumulative task budgets.
Comparisons use matched question IDs, and results are averaged over questions.

ResearchSpec coverage follows \cref{app:researchspec-coverage-evaluation};
the comparison reuses the existing coverage judgments.

\paragraph{Explicit and implicit requirements.}
We group the original ResearchRubrics criteria by their supplied
\emph{Explicit Criteria} and \emph{Implicit Criteria} labels. For each
question and category, the score is 100 times the sum of satisfied
criteria's signed weights divided by the sum of positive weights in that
category. The table reports the mean of these per-question scores; both
categories are present throughout the development subset. Negative-weight criteria
retain their penalties. Total uses all rubric categories and their full
positive-weight denominator, so it is not the average of the two displayed
subscores. Category scores are computed from the same Gemini 2.5 Pro
criterion-level verdicts as the total score.
Rubrics and diagnostic judgments are never supplied to the generating
agents.

\subsection{Editor Refinement: Settings and Paired Results}
\label{app:editor-refinement-protocol}
\label{app:editor-longcat}

LongCat uses fixed development subsets for each benchmark and five
conditions. Thinking is enabled and seed 42
is requested. The development subsets were expanded after inspection of the initial results; this is exploratory
rather than independent confirmation.

A0 is the unedited draft. A1 is the result of one original-Editor pass on
A0. The enhanced-Editor sequence is A0 $\rightarrow$ B1 $\rightarrow$ B2
$\rightarrow$ B3, where B1, B2, and B3 denote one, two, and three rounds.
A1 is a separate control and is not used as input to this sequence. The enhanced Global Editor receives citation-bound
excerpts from already retrieved sources, limited to 24,000 characters in
total and 2,400 per source, without new retrieval. Source access,
fact-checking instructions and iteration count form a joint intervention.
Per-call output permits 64,000 tokens. Cumulative task limits are
600 model calls, 1,048,576 completion tokens, 500 searches, and 500 page reads. Failed stages retain original costs and accepted artifacts.
LongCat recovery includes directive-bound URL correction and syntax repair
in auxiliary quotation fields, preserving native verdicts and scores.

The native scoring rules follow the shared conventions above. Each condition has one completed
grade; official reference-result alignment remains unestablished.

\paragraph{Readability results.}
Readability preference follows \cref{app:readability-preference-evaluation}.
The aggregate trend need not hold within each benchmark. For example,
LongCat B3 has mean margins of $-0.183$ on DRB-II and $+0.500$ on
ResearchRubrics relative to A0, yielding $+0.158$ overall.

\paragraph{Paired results.}
\Cref{tab:editor-longcat-native-paired} reports the paired contrasts supporting
\cref{tab:editor-refinement-native}. Intervals use 10,000 question-bootstrap
draws with seed 42 and are unadjusted for multiple comparisons.
LongCat B3--A0 changes DRB-II by $+2.51$ and
ResearchRubrics by $+0.09$. These intervals cover question variation,
not repeated-generation or repeated-Judge variance.

\begin{table}[!ht]
\centering\small
\setlength{\tabcolsep}{5pt}
\begin{tabular}{llrl}
\toprule
Benchmark & Difference & Mean change & Descriptive 95\% interval \\
\midrule
DRB-II & A1 $-$ A0 & $+0.17$ & $[-0.71,\ +1.11]$ \\
DRB-II & B3 $-$ A0 & $+2.51$ & $[+0.63,\ +4.78]$ \\
DRB-II & B1 $-$ A1 & $-0.89$ & $[-2.51,\ +0.56]$ \\
DRB-II & B2 $-$ B1 & $+0.52$ & $[-0.57,\ +1.70]$ \\
DRB-II & B3 $-$ B2 & $+2.70$ & $[+0.51,\ +5.24]$ \\
ResearchRubrics & A1 $-$ A0 & $-0.25$ & $[-2.08,\ +1.89]$ \\
ResearchRubrics & B3 $-$ A0 & $+0.09$ & $[-0.79,\ +1.18]$ \\
ResearchRubrics & B1 $-$ A1 & $-0.48$ & $[-3.00,\ +1.89]$ \\
ResearchRubrics & B2 $-$ B1 & $+1.69$ & $[-0.03,\ +3.59]$ \\
ResearchRubrics & B3 $-$ B2 & $-0.87$ & $[-2.35,\ +0.58]$ \\
\bottomrule
\end{tabular}
\caption{LongCat paired native-score changes on fixed development subsets.
A0 is the unedited draft; A1 applies the original Editor to A0 once.
B1--B3 are successive enhanced-Editor rounds starting from A0, independently
of A1. Each difference subtracts the second condition from the first;
positive changes favor the first condition.
DRB-II denotes DeepResearchBench II.}
\label{tab:editor-longcat-native-paired}
\end{table}

\paragraph{What the edits change.}
On DRB-II 58, editing corrects the online-learning study table from
``Arab Region'' and 3,348 participants to ``Jordan'' and
``280 students + 50 faculty,'' consistent with the detailed discussion.
The native score rises by 1.59 points. On the universal basic income task, clearer pilot-specific
sections gain a structural criterion, but weaker attribution of quantitative
claims loses a higher-weight citation criterion, yielding $-1.25$ points.
These examples show a factual correction in one report and a loss of
attribution quality in another.

Some score changes reflect Judge sensitivity: most newly credited
hydroforming facts were already in the unedited draft (A0). A higher score
therefore need not indicate that editing added the credited information. No model parameters are updated in this study.

\FloatBarrier
\section{Case Studies: From ResearchSpec to Edited Report}
\label{app:case-studies}
\lcmeta{English-language examples from the selected DeepResearchBench II cohort.
Evaluated with GPT-5.5 (medium).}
\lcmeta{In the excerpts, S$x.y$ identifies subsection $y$ of section $x$;
HP1 labels the Critic's first listed issue. ReportSpec is the name used
for ResearchSpec in the quoted source artifact. MPT denotes modern portfolio
theory, CPPI constant-proportion portfolio insurance, and CI confidence interval.}

\begin{lccase}{\textbf{CASE 01}\quad AI-enhanced portfolio management}
\phantomsection\label{app:case-high}
\lcmeta{DeepResearchBench II 012\hfill FINAL SCORE\;\textcolor{lcCaseDark}{\textbf{89.29 / 100}}}
\begin{lcband}{TASK \& SCOPE}
Review classical theories, AI/ML applications, multi-criteria decision making, and portfolio optimization and rebalancing through April 2024.
\lcnote{Task synopsis. The complete ResearchSpec assigns 19 subsections.}\end{lcband}
\begin{lcband}{RESEARCHSPEC}
\lcmeta{research\_spec.md\quad /\quad ReportSpec\quad /\quad selected original passages}
\lcfield{Global research boundaries}
\lcquote{\textbf{Output language}: English (matching the research question).}
\lcomit\lcfield{S1.2 | Capital Asset Pricing Model (CAPM)}
\lcmeta{Within S1: Classic Theoretical Frameworks}\lcfield{What to Cover}
\lcquote{Present CAPM as the equilibrium asset pricing model building on MPT. Explain the core idea: expected excess return of any asset is proportional to its beta (systematic risk) relative to the market portfolio.}
\lcomit\lcfield{Research Questions}
\lcquote{Who originated CAPM and in what years (Sharpe 1964, Lintner 1965, Mossin 1966)?}
\lcomit\lcfield{Required Entities / Cases}
\lcomit\lcquote{Jan Mossin (1966)}
\lcquote{Security Market Line (SML)}
\lcomit\lcfield{Source Leads}
\lcquote{https://www.investopedia.com/terms/c/capm.asp — CAPM overview (orientation only; find primary peer-reviewed source)}
\lcomit\end{lcband}
\begin{lcband}[neutral]{FINAL EVALUATION}
\lcnote{The plan still omits Treynor (1962); the final report also misses CPPI and fails the chronological-order criterion.}\end{lcband}
\end{lccase}
\lcmeta{Source text is abridged, not rewritten. Field labels and Markdown are typeset for readability;
[\ldots] marks omitted passages. The panels show selected ResearchSpec fields and source passages.}
\begin{lccase}{\textbf{CASE 01}\quad From the plan to the edited report}
\lcmeta{Recorded stage changes\hfill DeepResearchBench II 012}
\begin{lcband}{01\quad PLANNING REFINEMENT}
\lcfield{Critic identifies a missing method}
\lcquote{HP1: Heaton, Polson \& Witte (2017) — "Deep Portfolio Theory"}
\lcfield{Reviser adds a research question in S2.2}
\lcquote{How does Heaton, Polson \& Witte (2017) "Deep Portfolio Theory" propose learning portfolio weights directly from data, and how does this differ from the predict-then-optimize approach? \emph{(HP1)}}
\lcnote{Heaton et al. is absent from the merged spec and present in the revised spec. This changes the research agenda before section writing.}\end{lcband}
\begin{lcband}[neutral]{02\quad RESEARCH DRAFT}
\lcmeta{S1.2 / CAPM researcher / excerpt before editing}
\lcquote{Using the Fama-MacBeth methodology on NYSE, AMEX, and NASDAQ stocks (1963–1990), they found that the relation between market beta and average return was \textbf{flat} — beta had no explanatory power for the cross-section of average returns when used alone.}
\end{lcband}
\begin{lcband}{03\quad GLOBAL EDIT}
\lcmeta{Ownership decision after the section drafts are assembled}
\lcquote{Fama-French 1992 empirical findings {\fontsize{10}{12}\selectfont\lcarrow{}} S1.5; S1.2 cross-references S1.5}
\end{lcband}
\begin{lcband}{04\quad LOCAL EDIT \lcarrow{} FINAL REPORT}
\lcmeta{S1.2 / revised passage}
\lcquote{\emph{Fama-French (1992).} The most damaging empirical challenge to the CAPM came from Fama and French's (1992) finding that market beta had no explanatory power for the cross-section of average returns, while size and book-to-market equity were strongly and robustly related to average returns (discussed in detail in S1.5).}
\lcnote{This passage survives in the scored final report. The CAPM discussion keeps the principal finding and points to S1.5 for the detailed treatment.}\end{lcband}
\begin{lcband}[neutral]{WHAT CHANGED}
The plan gains a specific research question. Section research expands assignments into cited prose. Editing coordinates repeated material across sections through an ownership decision and a shorter cross-reference.\lcfield{What this does not establish}Only the final report has a benchmark score. This trace does not measure the score contribution of planning or editing.\end{lcband}
\end{lccase}
\lcmeta{Quotes retain the recorded wording. Commentary describes observed text changes, not measured intermediate score gains.}
\begin{lccase}{\textbf{CASE 02}\quad Generative AI and the labor market}
\phantomsection\label{app:case-low}
\lcmeta{DeepResearchBench II 056\hfill FINAL SCORE\;\textcolor{lcCaseDark}{\textbf{11.43 / 100}}}
\begin{lcband}{TASK AND RESEARCH AGENDA}
Review pre-June 2023 literature on positive views, negative views,
challenges and opportunities, followed by a study-level application table.
The ResearchSpec contains fourteen subsections across these themes and a
separate summary-table section.
\end{lcband}
\begin{lcband}{CRITIC \lcarrow{} REVISED SPEC}
\lcquote{HP1: Peng et al. (2023) --- GitHub Copilot Developer Productivity Study}
\lcnote{The merged Spec does not name Peng et al. or GitHub Copilot.
The revised Spec adds the study, its arXiv source lead, and a research question:}
\lcquote{What productivity effect did Peng et al. (2023) document for software developers using GitHub Copilot, and what does the heterogeneous-effects finding suggest about career transitions into software development? [HP1]}
\end{lcband}
\begin{lcband}{RESEARCH \lcarrow{} FINAL REPORT}
\lcquote{The treatment group completed the task \textbf{55.8\% faster} than the control group (71.17 minutes vs. 160.89 minutes on average, p = 0.0017, 95\% CI: 21--89\%).}
\lcnote{This sentence appears in the Researcher output and survives editing
in the scored final report. Global editing also assigns the aggregate
labor-exposure figures to S2.1, and the Local Editor applies that cross-reference.}
\end{lcband}
\begin{lcband}[neutral]{REMAINING COVERAGE GAP}
The 121,366-character final report emphasizes aggregate labor-market
exposure and prominent productivity studies. It omits the rubric-required
studies \emph{Can AI help for scientific writing?} and
\emph{Comparing physician and AI chatbot responses to patient questions
posted to a public social media forum}. Both titles are absent from the
revised Spec and final report, and information recall remains zero.
\end{lcband}
\begin{lcband}{TAKEAWAY}
The planning addition and its downstream passage are traceable, while the
remaining coverage gap persists. This case illustrates how refining an
existing agenda can leave other required evidence uninvestigated; its final
score does not isolate the contribution of an individual stage.
\end{lcband}
\end{lccase}
\lcmeta{Quotes are abridged from the planning, research, and editing outputs.}

\clearpage
\section{Full Author List}
\label{app:author-list}

This work is authored by the following members of Meituan LongCat Team:

\begin{center}
\renewcommand{\arraystretch}{1.45}
\begin{tabularx}{\textwidth}{@{}*{4}{>{\centering\arraybackslash}X}@{}}
He Zhu\textsuperscript{*} & Yue Xu\textsuperscript{*} & Wanli Wu & Haolin Ren \\
Yuxin Bian & Jiarui Zhao & Rongzhi Zhang & Quanchi Weng \\
Jinghao Cui & Yu Fan & Yuhan Liu & Yunhu Ye \\
Jiyuan Ren & Fengcheng Yuan & Zhao Yang & Jiacheng Zhang \\
Yuchuan Dai & Ruixuan Xiao & Haozhe Sun & Xiangyuan Liu \\
Cheng Sun & Yao Du & Yiming Hao & Hongbo Guo \\
Shuo He & Lei Wang & Xunliang Cai & Yan Chen\textsuperscript{\textdagger} \\
Fan Yang\textsuperscript{\textdagger} & Lingchuan Liu\textsuperscript{\textdagger} & & \\
\end{tabularx}

\vspace{0.75em}

\textsuperscript{*}\textit{Equal contribution.}\qquad
\textsuperscript{\textdagger}\textit{Corresponding authors.}
\end{center}

\begin{center}
\small\href{mailto:longcat-team@meituan.com}{\texttt{longcat-team@meituan.com}}
\end{center}

\end{document}